\documentclass[10pt,twocolumn,letterpaper]{article}
\usepackage[pagenumbers]{cvpr}
\usepackage{graphicx,amsmath,amssymb,booktabs,multirow,xcolor}
\ifdefined\XeTeXversion
  \DeclareFontFamily{TU}{ptm}{}
  \DeclareFontShape{TU}{ptm}{m}{n}{<-> "[fonts/LiberationSerif-Regular.ttf]:mapping=tex-text"}{}
  \DeclareFontShape{TU}{ptm}{b}{n}{<-> "[fonts/LiberationSerif-Bold.ttf]:mapping=tex-text"}{}
  \DeclareFontShape{TU}{ptm}{m}{it}{<-> "[fonts/LiberationSerif-Italic.ttf]:mapping=tex-text"}{}
  \DeclareFontShape{TU}{ptm}{b}{it}{<-> "[fonts/LiberationSerif-BoldItalic.ttf]:mapping=tex-text"}{}
  \DeclareFontShape{TU}{ptm}{bx}{n}{<-> ssub*ptm/b/n}{}
  \DeclareFontShape{TU}{ptm}{bx}{it}{<-> ssub*ptm/b/it}{}
\fi
\usepackage{xspace}
\usepackage[pagebackref,breaklinks,colorlinks,allcolors=blue]{hyperref}
\usepackage[capitalize]{cleveref}
\newcommand{\ours}{Instant~LiDAR\xspace}

\usepackage{colortbl,tabularx}
\newcolumntype{R}{>{\raggedleft\arraybackslash}X}
\definecolor{oursblue}{HTML}{267EAA}
\definecolor{oursbg}{HTML}{EEF5F8}
\definecolor{refgray}{HTML}{F3F5F6}

\newcommand{\resultPAICD}{0.275}
\newcommand{\resultPAIGtwoGain}{23}
\newcommand{\resultPAISeconds}{127}
\newcommand{\resultWaymoCD}{0.276}
\newcommand{\resultWaymoGtwoGain}{13}
\newcommand{\resultWaymoSeconds}{66}

\hypersetup{
  pdftitle={Instant LiDAR: Feed-Forward Gaussian Reconstruction for LiDAR Simulation},
  pdfauthor={Xin Kang, Jiahui Huang, Jiawei Ren, Sipeng Zhang, Yu Chen, Weihua Zhang, Junjie Lai}
}

\begin{document}
\title{Instant~LiDAR: Feed-Forward Gaussian Reconstruction for LiDAR Simulation}
\author{%
  \begin{tabular}{@{}c@{}}
    Xin Kang, ~Jiahui Huang,
    ~Jiawei Ren, ~Sipeng Zhang, 
    ~Yu Chen, ~Weihua Zhang,
    ~Junjie Lai \\
    NVIDIA
  \end{tabular}%
}
\twocolumn[{%
\renewcommand\twocolumn[1][]{#1}%
\maketitle
\noindent\begin{minipage}{\textwidth}
\centering
\includegraphics[width=\textwidth]{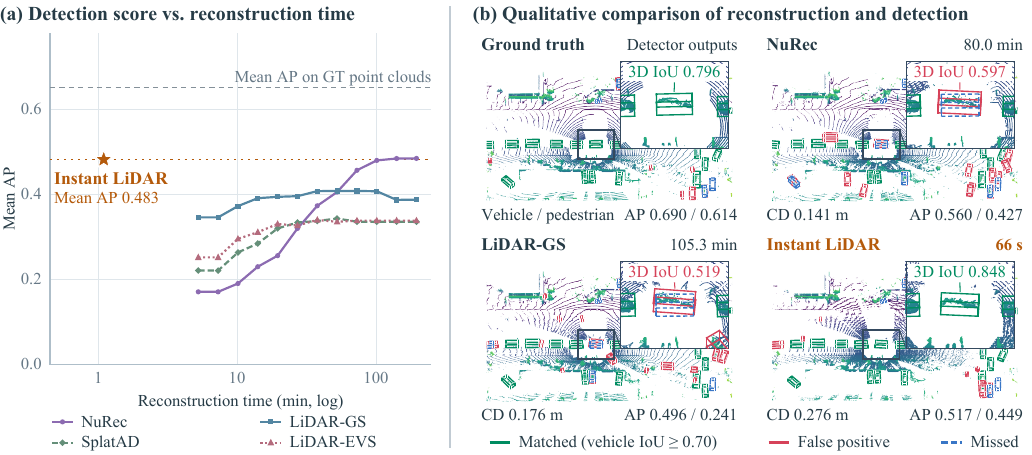}
\captionsetup{hypcap=false}
\captionof{figure}{\textbf{LiDAR reconstruction and detection on Waymo.}
(a) \ours{} reconstructs in 66\,s, \(\sim70\times\) faster than NuRec, with detection AP comparable with state-of-the-art per-scene optimization methods. (b) \ours{} preserves vehicle geometry and localization, enabling accurate object detection. CD and AP are averaged metrics on Waymo, and IoU refers to the highlighted vehicle.}
\label{fig:teaser}
\end{minipage}
\vspace{5pt}

}]
\begin{abstract}
LiDAR reconstruction supports closed-loop autonomous driving simulation, but hours of per-scene optimization make large asset libraries costly to build and maintain. We present \ours{}, a feed-forward pipeline that reconstructs multiple LiDAR scans as reusable Gaussian assets in minutes without per-scene optimization. LiDAR measurements provide strong geometric priors, but directly initializing Gaussians from them yields poor reconstruction due to nonuniform sampling and a limited field of view. Sparse returns leave surface gaps, while combining scans expands coverage but introduces redundant primitives. To address these challenges, \ours{} combines measurement-anchored Gaussian prediction, sampling-aware Gaussian scales, and multi-frame consolidation. The predictor refines attributes anchored to recorded returns, while the scales adapt spatial support to LiDAR sampling. Multi-frame consolidation merges overlapping predictions into a compact asset spanning the driving log. Compared with NuRec, \ours{} reconstructs scenes approximately $46\times$ faster on the Physical AI dataset (PAI) and $70\times$ faster on Waymo. It also achieves comparable downstream 3D detection accuracy on the Waymo dataset.
\end{abstract}

\section{Introduction}\label{sec:introduction}
LiDAR reconstruction provides reusable assets for closed-loop autonomous driving simulation~\cite{vista2,unisim}. 3D Gaussian Splatting (3DGS) based methods render LiDAR returns efficiently~\cite{nurec,lidargs,lidargspp,gslidar,lidarevs,splatad,lidarrt}, yet reconstructing a driving clip can still require hours of optimization~\cite{splatad,lidarrt}. Automotive developers need large asset libraries covering varied roads and traffic, with updates as new logs and reconstruction models become available~\cite{nurecscaling}. Repeated optimization makes reconstruction cost grow with library size and update frequency, motivating faster LiDAR reconstruction.

Feed-forward 3DGS methods reduce image-based scene reconstruction time from hours to seconds without per-scene optimization~\cite{drivingforward,storm}. A trained predictor processes new logs directly and is already used in industrial workflows such as NuRec~\cite{nurecscaling}. However, Gaussian-based LiDAR reconstruction still largely relies on per-scene optimization. A feed-forward method is therefore needed to reconstruct LiDAR scenes quickly while maintaining reconstruction quality and downstream detection accuracy.

Unlike RGB images, LiDAR can provide recorded 3D positions as strong geometric priors for Gaussian initialization~\cite{nfl,lidargs}. These priors alone, however, are insufficient for complete scene reconstruction. Nonuniform sampling leaves distant surfaces sparsely observed, so initialized Gaussians can leave gaps that appear as missing returns in rendered scans. Each frame's limited field of view also leaves parts of the scene unobserved. Aggregating per-frame Gaussians expands coverage, but redundant predictions in overlapping regions can bias rendered return depths.

To address these challenges, we propose \ours{} for feed-forward reconstruction of reusable Gaussian assets in minutes to achieve high-quality LiDAR rendering. It has three components: measurement-anchored Gaussian prediction, sampling-aware Gaussian scales, and multi-frame consolidation. The predictor learns attributes for Gaussians anchored to recorded LiDAR returns, and the scales adapt spatial support to LiDAR sampling. Multi-frame consolidation aligns per-frame predictions and merges overlapping static Gaussians by moment matching~\cite{runnalls}, reducing redundancy while preserving spatial extent.

Experiments on the Physical AI dataset (PAI) and Waymo~\cite{waymo} show approximately $46\times$ and $70\times$ faster scene reconstruction than NuRec, respectively. On Waymo, \ours{} also achieves comparable downstream 3D detection quality, as shown in \cref{fig:teaser}. With the sensor shifted 3\,m to the left, \ours{} outperforms all evaluated per-scene optimization methods in detection scores.

Our contributions are as follows:
\begin{itemize}
    \item We present \ours{}, a feed-forward LiDAR reconstruction pipeline that builds reusable Gaussian scene assets for LiDAR simulation in minutes.
    \item We combine sampling-aware Gaussian scales and multi-frame consolidation to address sparse surface coverage and redundant observations, improving coverage while compacting scene assets.
    \item We achieve approximately $46\times$ and $70\times$ faster scene reconstruction than NuRec on PAI and Waymo, respectively, while maintaining competitive downstream 3D detection quality on Waymo.
\end{itemize}

\begin{figure*}[t]\centering
\includegraphics[width=\textwidth]{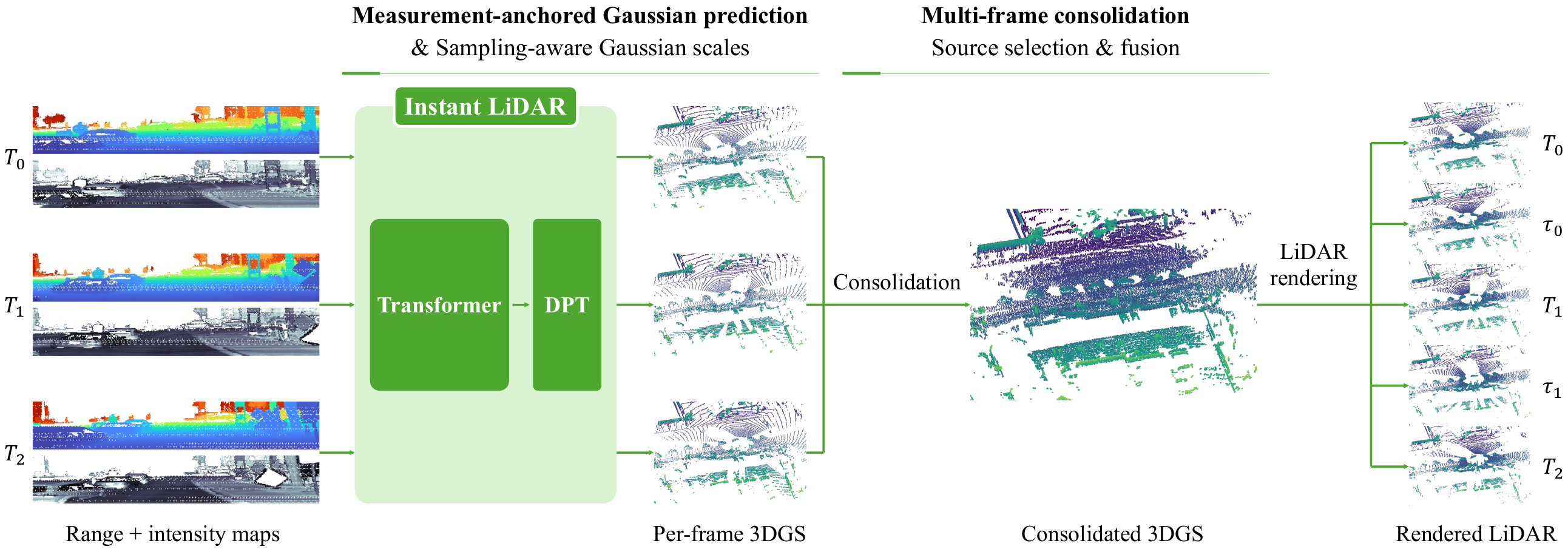}
\caption{\textbf{\ours{} overview.}
Range and intensity maps guide measurement-anchored Gaussian prediction and sampling-aware Gaussian scales (\cref{fig:anchor-support}). Multi-frame consolidation (\cref{fig:consolidation}) assembles a scene for LiDAR rendering at input and held-out frames ($\tau_0,\tau_1$).}
\label{fig:method}
\end{figure*}

\section{Related Work}\label{sec:related}
\subsection{LiDAR reconstruction}\label{sec:related-lidar-reconstruction}
LiDAR reconstruction recovers geometry and return attributes from LiDAR scans for LiDAR rendering. \mbox{LiDARsim}~\cite{lidarsim} and PCGen~\cite{pcgen} combine reconstructed geometry with learned sensor models. Neural fields model geometry and returns, with extensions to dynamic scenes and joint camera--LiDAR synthesis~\cite{lidarnerf,nfl,lidar4d,dynfl,unisim,neurad,alignmif}. NFL~\cite{nfl} incorporates beam divergence, secondary returns, and ray dropping into its sensor model. LiDAR-GS~\cite{lidargs}, GS-LiDAR~\cite{gslidar}, SplatAD~\cite{splatad}, and LiDAR-RT~\cite{lidarrt} accelerate rendering using explicit Gaussian primitives. LiDAR-GS++~\cite{lidargspp} and LiDAR-EVS~\cite{lidarevs} improve extrapolation through diffusion priors and pseudo-LiDAR supervision, respectively. These approaches generally require per-scene optimization, as does the standard workflow of NuRec~\cite{nurec}. Reconstruction can take hours~\cite{splatad,lidarrt}, making large scene libraries costly to build and update.

\subsection{Feed-forward 3DGS reconstruction}\label{sec:related-feedforward}
Feed-forward 3DGS methods predict Gaussian scenes with pretrained networks, avoiding optimization for each new scene. pixelSplat~\cite{pixelsplat} predicts Gaussians from image pairs. MVSplat~\cite{mvsplat} uses a plane-sweep cost volume to estimate depth and locate Gaussian centers, while DepthSplat~\cite{depthsplat} incorporates pretrained monocular depth features into multi-view depth estimation. DrivingForward~\cite{drivingforward} supports surround-view inputs, while STORM~\cite{storm} and EVolSplat~\cite{evolsplat} aggregate observations across frames. Learning Gaussian Structure~\cite{lgs} uses LiDAR-initialized primitives but targets image reconstruction. LiDAR simulation instead requires range and intensity returns under the sensor's angular sampling pattern. Our focus is feed-forward reconstruction of Gaussian scenes for LiDAR simulation, using recorded scans to produce reusable assets in minutes while preserving the geometric structures needed for downstream perception.

\subsection{LiDAR generation}\label{sec:related-generation}
LiDAR generation learns a distribution of sensor observations to synthesize new scans. LiDARGen~\cite{lidargen} uses score-based denoising in a panoramic representation. Sensor2Sensor~\cite{sensor2sensor} generates camera--LiDAR logs from monocular videos, while Cosmos-Transfer-LidarGen~\cite{cosmoslidargen} generates LiDAR range maps from multi-view RGB images. These models produce sensor observations rather than reusable scene assets. We instead reconstruct Gaussian assets from recorded LiDAR scans for repeated rendering.

\section{Method}\label{sec:method}
Let $\mathcal X=\{(\mathbf R_t,\mathbf I_t,\mathbf M_t,\mathbf T_t)\}_{t=1}^{T}$ denote the input sequence of range maps, intensity maps, return-validity masks, and sensor poses. 
Given sensor calibration and actor tracks, \ours{} reconstructs a scene using a Gaussian representation~\cite{kerbl3Dgaussians}:
\begin{equation}
 \mathcal G=F_\theta(\mathcal X)
 =\{g_j\}_{j=1}^{N_G},
 \label{eq:scene-reconstruction}
\end{equation}
where $g_j=(\boldsymbol\mu_j,\boldsymbol\Sigma_j,\alpha_j,\rho_j)$ stores its center, covariance, opacity, and intensity. The pipeline shown in \cref{fig:method} comprises measurement-anchored Gaussian prediction (\cref{sec:prediction}), sampling-aware Gaussian scales (\cref{sec:support}), and multi-frame consolidation (\cref{sec:assembly}).

\subsection{Measurement-anchored Gaussian prediction}\label{sec:prediction}
We use recorded LiDAR positions and intensities as anchors and predict local attribute residuals, following earlier Gaussian representations~\cite{scaffoldgs,lidargs}. An eight-block Transformer~\cite{attention} encodes patches of range and intensity maps. It combines within-frame and cross-frame attention. A DPT~\cite{dpt} head is used to fuse features from four encoder layers and upsample them to predict Gaussian attributes at the input resolution.

\begin{figure}[t]
\centering
\includegraphics[width=\linewidth]{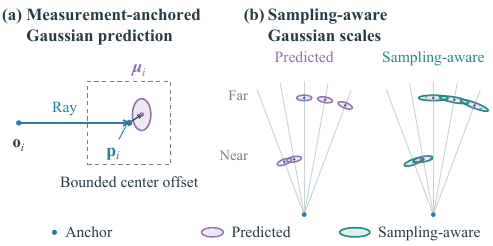}
\caption{\textbf{Measurement-anchored Gaussian prediction and sampling-aware Gaussian scales.} (a) Bounded center offsets from recorded returns. (b) Sampling-dependent support grows with range, shown in the beam frame before learned rotation.}
\label{fig:anchor-support}
\end{figure}

Specifically, for each input frame, ray $i$ has origin $\mathbf o_i$, unit direction $\mathbf d_i$, recorded range $r_i$, and intensity $I_i$. The return position is $\mathbf p_i=\mathbf o_i+r_i\mathbf d_i$. Each valid return anchors a Gaussian (\cref{fig:anchor-support}(a)) whose center and intensity are
\begin{align}
 \boldsymbol\mu_{i}&=\mathbf p_i+\delta_{\max}\tanh(\boldsymbol\delta_{i}),\label{eq:center}\\
 \rho_{i}&=I_i+\epsilon_{\max}\tanh(\eta_{i}),\label{eq:intensity}
\end{align}
where the network predicts the position and intensity residuals $\boldsymbol\delta_{i}$ and $\eta_{i}$. 
The bounds $\delta_{\max}$ and $\epsilon_{\max}$ limit position offsets per coordinate and intensity residuals, respectively.

The network also predicts positive scales, opacities, and a rotation residual relative to the LiDAR beam frame.

\subsection{Sampling-aware Gaussian scales}\label{sec:support}
LiDAR rays spread farther apart with distance, leaving gaps between small Gaussians. Inspired by sampling-aware rendering~\cite{ewa,mipnerf,mipsplatting}, we adapt local scales to the ray spacing.

For source-to-center range $R_i=\|\boldsymbol\mu_i-\mathbf o_i\|$, angular intervals $\Delta_{a,i}$ in radians, $a\in\{\mathrm{az},\mathrm{el}\}$, give spatial intervals of approximately $R_i\Delta_{a,i}$. We obtain these angles from neighboring azimuth columns and sorted beam elevations, and define
\begin{equation}
 b_{a,i}=k_{a,i}R_{i}\Delta_{a,i}.
 \label{eq:sampling-scale}
\end{equation}
The network predicts $\kappa_{a,i}$, giving $k_{a,i}=0.3(1+0.5\tanh\kappa_{a,i})\in[0.15,0.45]$. Separate terms account for unequal azimuth and elevation spacing.

For predicted local scales $\widetilde s_{a,i}$ and $\widetilde s_{\mathrm{rad},i}$, we apply the variance addition used in Mip-Splatting~\cite{mipsplatting}:
\begin{equation}
 \begin{aligned}
 s_{a,i}&=\sqrt{\widetilde s_{a,i}^{2}+b_{a,i}^{2}},\\
 s_{\mathrm{rad},i}&=\sqrt{\widetilde s_{\mathrm{rad},i}^{2}+b_{\mathrm{rad}}^{2}}.
 \end{aligned}
 \label{eq:support}
\end{equation}
The terms $b_{a,i}$ set minimum Gaussian scales $s_{a,i}\ge b_{a,i}$ based on LiDAR ray spacing. \Cref{fig:anchor-support}(b) illustrates how wider ray spacing increases these scale floors. The fixed floor $b_{\mathrm{rad}}>0$ prevents the third local scale from collapsing to zero thickness during training.

After scale adjustment, we apply the final orientation $\mathbf Q_i$ to obtain
\begin{equation}
 \boldsymbol\Sigma_i=\mathbf Q_i\operatorname{diag}
 (s_{\mathrm{az},i}^2,s_{\mathrm{el},i}^2,s_{\mathrm{rad},i}^2)\mathbf Q_i^\top.
 \label{eq:covariance-decomposition}
\end{equation}
Here, $\mathbf Q_i=\mathbf Q_i^{\mathrm{beam}}\Delta\mathbf Q_i$, where $\mathbf Q_i^{\mathrm{beam}}$ defines the initial beam frame and $\Delta\mathbf Q_i$ is the learned rotation residual. The beam frame's axes follow the azimuth tangent, elevation tangent, and input ray direction.

\subsection{Multi-frame consolidation}\label{sec:assembly}
\begin{figure}[t]
\centering
\includegraphics[width=\linewidth]{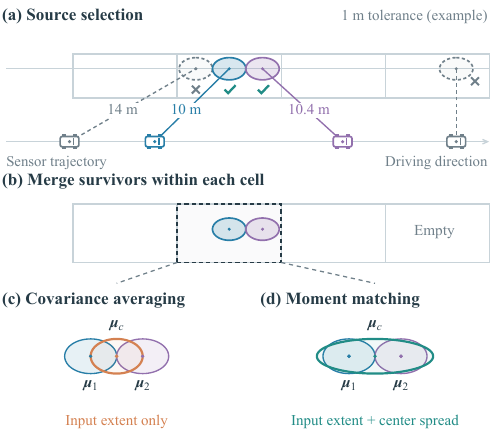}
\caption{\textbf{Multi-frame consolidation.} (a) Source distance and multi-window support select candidates within each cell. 
(b) Survivors merge within each cell. (c,d) Moment matching preserves center spread omitted by covariance averaging.} %
\label{fig:consolidation}
\end{figure}

Multi-frame consolidation addresses incomplete scan coverage and redundant predictions in overlapping frames. Dynamic Gaussians are assembled per actor using the provided tracks and time-dependent transforms~\cite{nsg,dynfl,streetgaussians}. Static Gaussians undergo source selection and moment-matching fusion as shown in \cref{fig:consolidation}.

For source selection, we register centers and covariances using the sensor poses $\{\mathbf T_t\}_{t=1}^{T}$ and partition them into cubic cells of width $h$. Each Gaussian's source window comprises the input frames used to predict it. We score each candidate by its distance to the nearest sensor position in that window. Candidates within a tolerance of the cell's minimum score survive, favoring closer observations. Cells observed in only one window are discarded (\cref{fig:consolidation}(a)).

The survivors in each cell form a group $c$, merged into one Gaussian by standard moment matching~\cite{runnalls} (\cref{fig:consolidation}(b)). With equal weights $\beta_j=1/|c|$ and $\Delta\boldsymbol\mu_j=\boldsymbol\mu_j-\boldsymbol\mu_c$, we preserve the group's mean and covariance:
\begin{align}
 \boldsymbol\mu_c&=\sum_{j\in c}\beta_j\boldsymbol\mu_j,\nonumber\\
 \boldsymbol\Sigma_c&=
 \underbrace{\sum_{j\in c}\beta_j\boldsymbol\Sigma_j}_{\text{input extent}}
 +\underbrace{\sum_{j\in c}\beta_j\Delta\boldsymbol\mu_j
 \Delta\boldsymbol\mu_j^\top}_{\text{center spread}}.
 \label{eq:fusion}
\end{align}
The two terms preserve input extent (\cref{sec:support}) and center spread, respectively. Covariance averaging omits center spread, narrowing the representative when centers differ (\cref{fig:consolidation}(c,d)). Orientation follows the resulting covariance; opacity, intensity, and other attributes use the same weights. 

After fusion, we remove isolated static Gaussians. For centers within 40\,m of any input sensor position, we require at least 15 static centers within a 0.5\,m radius, including the center itself. Centers outside this region are retained to preserve sparse distant surfaces.

\subsection{Rendering, training, and inference}\label{sec:training}
We render range and intensity along query LiDAR rays using a backend based on 3DGUT~\cite{3dgut}. For each ray, let $u_j$ be the compositing weight of Gaussian $j$, with contributing depth $d_j$ and intensity $\rho_j$. For positive accumulated opacity, we normalize the weights as $w_j=u_j/\sum_k u_k$ and compute
\begin{equation}
 \widehat r=\sum_j w_jd_j,\qquad
 \widehat I=\sum_j w_j\rho_j.
 \label{eq:rendered-returns}
\end{equation}

We train the encoder and prediction head with the reconstruction objective
\begin{equation}
 \mathcal L=\mathcal L_{\mathrm{range}}
             +\mathcal L_{\mathrm{intensity}}+0.1\mathcal L_{\mathrm{opacity}},
 \label{eq:training}
\end{equation}
where $\mathcal L_{\mathrm{range}}$ is the $L_1$ range error, $\mathcal L_{\mathrm{intensity}}$ is the intensity mean squared error, and $\mathcal L_{\mathrm{opacity}}$ encourages unit accumulated opacity on valid-return rays through an $L_1$ loss. Gaussians are instantiated only at valid input returns, so source-return validity is supplied by the input mask. We optimize the predictor with Adam~\cite{adam}.

\begin{table*}[!t]\centering
\caption{\textbf{Reconstruction quality and time.} Results average clip means over six clips per dataset. Distances are in meters; CD=p2G+G2p and F@5 is the F-score at 5\,cm. Bold/underline mark the best/second-best geometric scores. All methods are evaluated on the same NVIDIA H20 GPU, with reconstruction time measured from LiDAR input to 3DGS output.}\label{tab:main}
\fontsize{9}{10.4}\selectfont
\setlength{\tabcolsep}{3pt}
\renewcommand{\arraystretch}{1.12}
\begin{tabularx}{\textwidth}{@{}lRRRRRRRRRR@{}}
\toprule
 & \multicolumn{5}{c}{PAI} & \multicolumn{5}{c}{Waymo} \\
\cmidrule(lr){2-6}\cmidrule(lr){7-11}
Method & CD$\downarrow$ & p2G$\downarrow$ & G2p$\downarrow$ & F@5$\uparrow$ & Time (s) & CD$\downarrow$ & p2G$\downarrow$ & G2p$\downarrow$ & F@5$\uparrow$ & Time (s) \\
\midrule
Isotropic splats & 0.725 & 0.187 & 0.538 & 0.254 & 79 & 0.499 & 0.261 & 0.238 & 0.261 & 55 \\
\addlinespace[2pt]
NuRec & \textbf{0.229} & \underline{0.104} & \textbf{0.125} & \textbf{0.513} & 5875 & \textbf{0.141} & \textbf{0.076} & \textbf{0.065} & \textbf{0.641} & 4802 \\
LiDAR-GS & \underline{0.258} & 0.113 & \underline{0.145} & 0.455 & 5818 & \underline{0.176} & \underline{0.100} & \underline{0.076} & \underline{0.581} & 6315 \\
SplatAD & 0.336 & 0.140 & 0.196 & 0.352 & 4214 & 0.216 & 0.111 & 0.105 & 0.417 & 3318 \\
LiDAR-EVS & 0.356 & 0.157 & 0.200 & 0.340 & 4702 & 0.222 & 0.117 & 0.104 & 0.431 & 3547 \\
\addlinespace[2pt]
\rowcolor{oursbg} \textcolor{oursblue}{\textbf{\ours}} & 0.275 & \textbf{0.094} & 0.180 & \underline{0.495} & 127 & 0.276 & 0.148 & 0.128 & 0.466 & 66 \\
\bottomrule
\end{tabularx}
\end{table*}

At inference, measurement-anchored Gaussian prediction uses fixed network weights. Sampling-aware Gaussian scales and multi-frame consolidation yield a compact scene asset. When rendering point clouds, contributions from separated surfaces can produce a mean depth between them, resulting in floating points. Inspired by the depth distortion loss in 2DGS~\cite{2dgs}, we measure depth dispersion at inference using the weighted depth variance
\begin{equation}
 \sigma_d^2=\sum_j w_j(d_j-\widehat r)^2.
 \label{eq:dispersion}
\end{equation}
This post-processing filter removes returns with low accumulated opacity or high depth dispersion, producing cleaner rendered point clouds. 

\begingroup
\clubpenalty=10000
\section{Experiments}\label{sec:experiments}
In this section, we compare reconstruction quality and time (\cref{sec:reconstruction-evaluation}) and evaluate downstream detection along recorded and laterally shifted trajectories (\cref{sec:detection-evaluation}). Ablations examine learned attributes, sampling-aware Gaussian scales, and multi-frame consolidation (\cref{sec:ablation}).

\subsection{Evaluation protocol}
\paragraph{Datasets.}
Following the evaluation scale of prior LiDAR reconstruction studies~\cite{lidargs,lidarrt}, we select six clips each from PAI and Waymo~\cite{waymo} for evaluation. The non-public Physical AI (PAI) driving dataset uses a 128-beam LiDAR. We train on 723 clips and randomly select the evaluation clips from a disjoint test set, holding out every tenth frame to obtain 60 queries. We plan to release the data used in this study. Waymo uses the 64-beam TOP sensor. We exclude the eight evaluation segments selected by LiDAR-GS~\cite{lidargs} (v1) from Waymo's official training split, using 771 segments for training and 19 for validation. Six eligible evaluation segments follow its 50-source/four-query protocol, yielding 24 held-out queries disjoint from training and validation. The supplementary material details the splits.

\paragraph{Methods.}
We compare with four methods with publicly available implementations, including NuRec~\cite{nurec} (release 26.06), LiDAR-GS~\cite{lidargs}, SplatAD~\cite{splatad}, and LiDAR-EVS~\cite{lidarevs}. We fixed an unguarded division in LiDAR-GS's released 2DGS rasterizer; the unmodified code diverged in 13/13 attempted runs in our setup. All reported LiDAR-GS results use this fix. Three learning-free baselines are included: \emph{Isotropic splats} use fixed-size spherical Gaussians; \emph{Sampling-aware splats} set scales according to LiDAR sampling geometry; \emph{\ours{} w/o learning} uses fixed attribute rules and our multi-frame consolidation and post-processing filter.

\paragraph{Metrics and hardware.}
Reconstruction uses prediction-to-GT (p2G) and GT-to-prediction (G2p) mean nearest-neighbor distances, their sum (Chamfer distance, CD), and F-score~\cite{tanksandtemples} at 5\,cm (F@5). The two directions measure reconstruction accuracy and coverage, respectively. Ground-truth (GT) point clouds are recorded scans within a common $[3,100]$\,m radial range; frame scores are averaged within clips, then clips are weighted equally. 
Detection uses Waymo~\cite{waymo} average precision (AP) and AP weighted by heading accuracy (APH). All training and inference use NVIDIA H20 GPUs. We measure reconstruction time from input LiDAR scans to a renderable 3DGS asset on the same H20 GPU for all methods. 

\begin{figure*}[!t]\centering
\includegraphics[width=\textwidth]{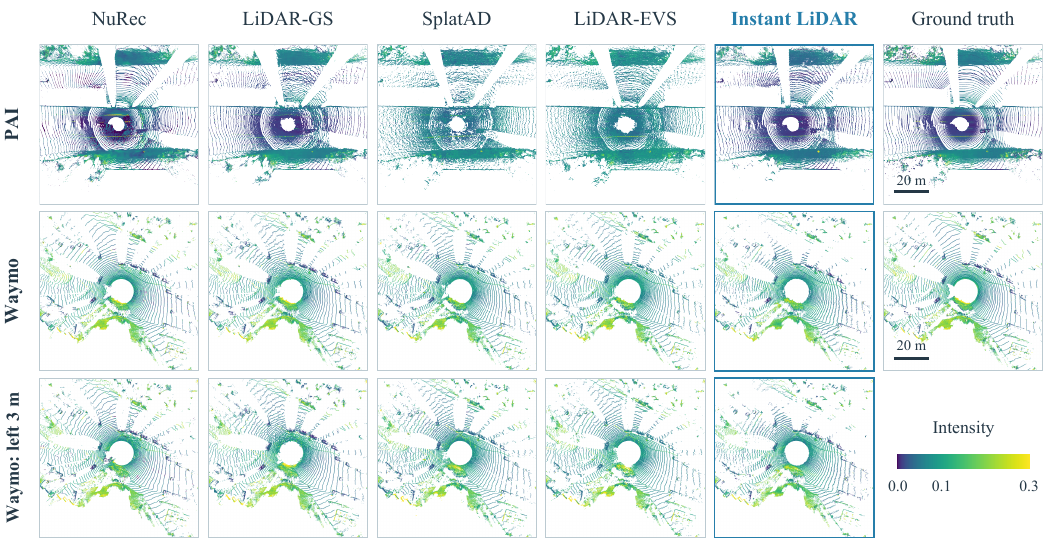}
\caption{\textbf{Reconstruction at recorded and extrapolated poses.} The first two rows show held-out PAI and Waymo queries; the third shifts the Waymo sensor 3\,m left. The paired Waymo views retain the same timestamp, heading, and coordinate frame. All panels use a $90\times90$\,m crop and the intensity scale at lower right. No GT scan is available at the shifted pose.}
\label{fig:reconstruction}
\end{figure*}

\begin{table*}[!t]\centering
\caption{\textbf{Waymo detection across query settings.} LEVEL\_1 scores pool six sequences per setting. Shifted queries use the same 314 timestamps at a 3\,m left offset, without GT scans. Bold/underline mark the best/second-best reconstructions.}\label{tab:detection}
\fontsize{9}{10.4}\selectfont
\setlength{\tabcolsep}{3pt}
\renewcommand{\arraystretch}{1.08}
\begin{tabularx}{\textwidth}{@{}l*{12}{R}@{}}
\toprule
 & \multicolumn{4}{c}{Held-out: 24 frames} & \multicolumn{4}{c}{Conditioning: 290 frames} & \multicolumn{4}{c}{Left-shifted 3\,m: 314 frames} \\
\cmidrule(lr){2-5}\cmidrule(lr){6-9}\cmidrule(lr){10-13}
 & \multicolumn{2}{c}{Vehicle} & \multicolumn{2}{c}{Pedestrian} & \multicolumn{2}{c}{Vehicle} & \multicolumn{2}{c}{Pedestrian} & \multicolumn{2}{c}{Vehicle} & \multicolumn{2}{c}{Pedestrian} \\
\cmidrule(lr){2-3}\cmidrule(lr){4-5}\cmidrule(lr){6-7}\cmidrule(lr){8-9}\cmidrule(lr){10-11}\cmidrule(lr){12-13}
Point cloud & AP$\uparrow$ & APH$\uparrow$ & AP$\uparrow$ & APH$\uparrow$ & AP$\uparrow$ & APH$\uparrow$ & AP$\uparrow$ & APH$\uparrow$ & AP$\uparrow$ & APH$\uparrow$ & AP$\uparrow$ & APH$\uparrow$ \\
\midrule
\rowcolor{refgray} GT point clouds & 0.671 & 0.665 & 0.591 & 0.503 & 0.692 & 0.684 & 0.616 & 0.527 & --- & --- & --- & --- \\
NuRec & 0.445 & 0.433 & 0.250 & 0.177 & \textbf{0.569} & \textbf{0.558} & \underline{0.443} & \underline{0.363} & 0.247 & 0.238 & 0.053 & 0.042 \\
LiDAR-GS & 0.439 & 0.432 & 0.154 & 0.141 & 0.501 & 0.493 & 0.249 & 0.214 & 0.341 & 0.331 & 0.120 & 0.097 \\
SplatAD & 0.438 & 0.426 & 0.225 & 0.163 & 0.430 & 0.421 & 0.226 & 0.170 & 0.258 & 0.251 & 0.110 & 0.080 \\
LiDAR-EVS & 0.419 & 0.411 & 0.270 & 0.189 & 0.399 & 0.391 & 0.264 & 0.206 & 0.357 & 0.349 & 0.212 & 0.162 \\
\ours{} w/o learning & \underline{0.461} & \underline{0.447} & \underline{0.300} & \underline{0.254} & 0.469 & 0.457 & 0.337 & 0.268 & \underline{0.421} & \underline{0.406} & \underline{0.249} & \underline{0.202} \\
\rowcolor{oursbg} \textcolor{oursblue}{\textbf{\ours}} & \textbf{0.524} & \textbf{0.508} & \textbf{0.336} & \textbf{0.264} & \underline{0.516} & \underline{0.506} & \textbf{0.458} & \textbf{0.374} & \textbf{0.434} & \textbf{0.419} & \textbf{0.266} & \textbf{0.216} \\
\bottomrule
\end{tabularx}
\end{table*}

\subsection{Reconstruction quality and time}\label{sec:reconstruction-evaluation}
In \cref{tab:main}, we report CD, accuracy (p2G), coverage (G2p), F@5, and reconstruction time. Across the evaluated clips, CD of \ours{} is \resultPAICD{}\,m on PAI and \resultWaymoCD{}\,m on Waymo. On PAI, \ours{} p2G of 0.094\,m is the lowest among evaluated methods, but G2p is higher than NuRec's (0.180 versus 0.125\,m), showing that the CD gap comes from coverage. On Waymo, both accuracy and coverage contribute to the gap.
We also compare reconstruction time in \cref{tab:main}. The reconstruction time of \ours{} averages \resultPAISeconds{}\,s per scene on PAI and \resultWaymoSeconds{}\,s on Waymo, compared with 5875\,s and 4802\,s for NuRec. This yields approximately $46\times$ and $70\times$ faster reconstruction, respectively, showing the efficiency advantage of \ours{}.

The qualitative results shown in \cref{fig:reconstruction} compares reconstructions with GT scans at held-out timestamps and shows reconstructions at laterally shifted poses.
In the first two rows, \ours{} preserves the accurate road surface and nearby vehicle outlines, with sparser coverage of distant scene structures. 
The bottom two rows of \cref{fig:reconstruction} compare the same Waymo scene at its recorded pose and 3\,m to the vehicle's left, with timestamp and heading fixed. 
\ours{} retains distinct road-surface scan patterns and nearby vehicle outlines.
We further evaluate downstream object detection to assess the usefulness of reconstructed scenes for autonomous-driving perception in \cref{sec:detection-evaluation}.

The results in \cref{tab:main} and \cref{fig:reconstruction} show that \ours{} reconstructs reusable assets in minutes while retaining road and nearby vehicle structures, reducing the cost of building and refreshing scene libraries.

\begin{figure*}[!t]\centering
\includegraphics[width=\textwidth]{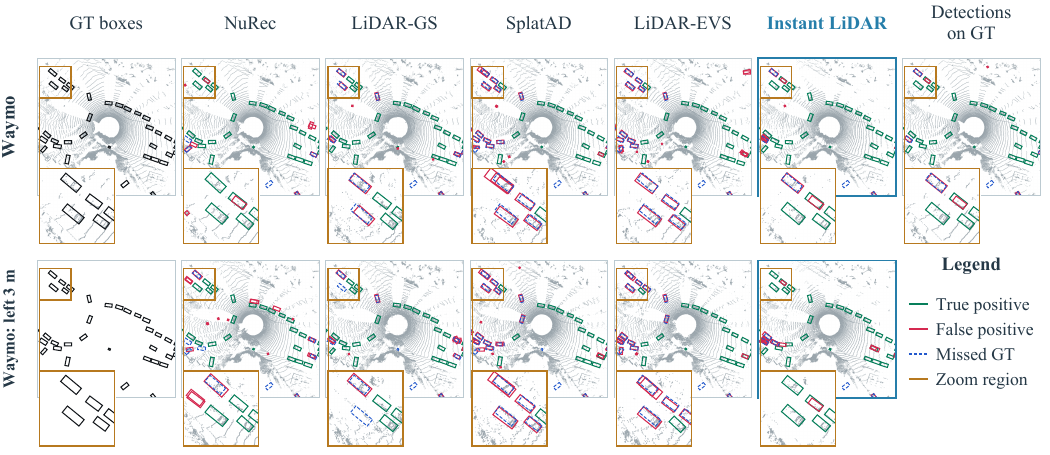}
\caption{\textbf{Detection at recorded and extrapolated poses.} The two rows evaluate the Waymo example from \cref{fig:reconstruction} at its recorded pose (top) and 3\,m to the left (bottom). Insets enlarge the same vehicle group. Columns compare GT boxes, detections on five reconstructions, and detections on GT point clouds. The shifted GT panel shows annotations without a GT point cloud.}
\label{fig:detection}
\end{figure*}

\begin{figure*}[!t]\centering
\includegraphics[width=\textwidth]{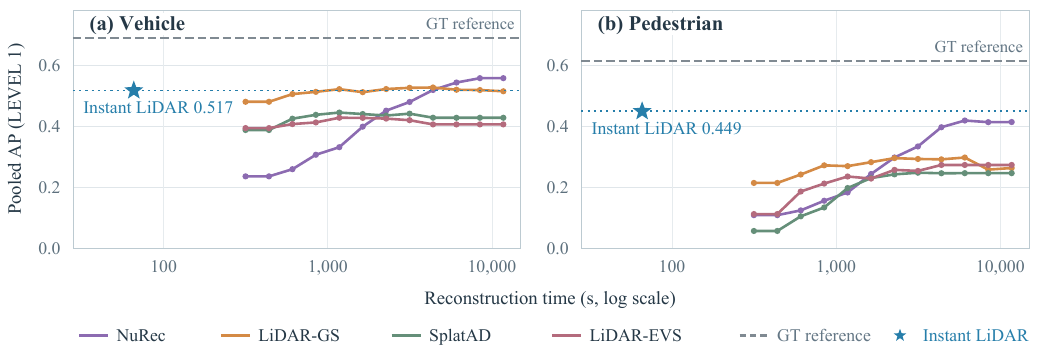}
\caption{\textbf{Detection quality versus reconstruction time.} AP is pooled over 290 conditioning and 24 held-out frames on the recorded trajectory. Curves show independent retraining runs; stars mark our mean reconstruction time and pooled AP. Dashed lines show AP on GT point clouds. Reconstruction time is measured from input to a renderable 3DGS asset; query rendering and detection are excluded.}
\label{fig:detection-time}
\end{figure*}

\subsection{Detection quality versus reconstruction time}\label{sec:detection-evaluation}
Following prior work~\cite{dynfl,lidargs}, we evaluate reconstruction quality through downstream object detection. We apply a frozen CenterPoint~\cite{centerpoint} detector to ground truth and reconstructed TOP first-return point clouds. We separately score 290 input (conditioning) and 24 held-out frames on the recorded trajectory. For extrapolation, we shift all 314 queries 3\,m left, retaining headings and actor states. This evaluates detection of the original annotated objects after the viewpoint change. All methods share the detector, timestamp-specific annotations, and coordinate frame.

As shown in \cref{tab:detection}, on held-out frames, \ours{} w/o learning exceeds all four per-scene baselines with vehicle/pedestrian AP of 0.461/0.300. Learning adds 0.063/0.036 AP, reaching 0.524/0.336 and the highest pooled AP/APH among evaluated reconstructions. On conditioning frames, \ours{} scores 0.516/0.458 versus NuRec's 0.569/0.443; NuRec's vehicle AP falls to 0.445 on held-out queries, reversing its lead. The top row of \cref{fig:detection} illustrates the held-out query at its recorded pose. \ours{} produces vehicle boxes with locations and orientations similar to detections on GT point clouds. The enlarged region also exposes missed or displaced boxes in other baseline reconstructions. 

As shown in \cref{tab:detection}, on the laterally shifted trajectory, \ours{} achieves vehicle/pedestrian AP of 0.434/0.266 and APH of 0.419/0.216, the highest pooled scores among the evaluated reconstructions. The strongest per-scene baseline, LiDAR-EVS, reaches AP of 0.357/0.212. Learning improves on the fixed-attribute variant by 0.013/0.017 AP. The bottom row of \cref{fig:detection} shows the shifted results. \ours{} retains correctly localized vehicle detections, whereas several baselines miss or misalign vehicles in the same group. Across all 314 queries, extrapolation remains harder than replay. These results support the use of our assets for lateral viewpoint changes. 

In \cref{fig:detection-time}, we compare pooled AP against scene reconstruction time on all 314 original-trajectory frames. \ours{} reaches vehicle/pedestrian AP of 0.517/0.449 in an average of \resultWaymoSeconds{}\,s per scene. Vehicle AP of \ours{} is close to LiDAR-GS at longer budgets, while NuRec requires several thousand seconds to exceed our score. Pedestrian AP of \ours{} exceeds all four baselines at every evaluated budget. These curves show that \ours{} offers competitive detection quality with minute-scale reconstruction.

\subsection{Ablation studies}\label{sec:ablation}
In this section, we ablate learned attributes, sampling-aware Gaussian scales, and multi-frame consolidation, including moment matching. The detailed results are shown in \cref{tab:ablation,tab:consolidation}.

\begin{table}[!htbp]\centering
\caption{\textbf{Learned attributes and sampling-aware Gaussian scales.} CD is in meters. \ours{} variants share consolidation and post-processing filter. The model without the sampling term is retrained.}\label{tab:ablation}
\fontsize{9}{10.4}\selectfont
\setlength{\tabcolsep}{3pt}
\renewcommand{\arraystretch}{1.12}
\begin{tabularx}{\columnwidth}{@{}lRR@{}}
\toprule
Variant & PAI$\downarrow$ & Waymo$\downarrow$ \\
\midrule
\multicolumn{3}{@{}l}{\emph{Initialization baselines}} \\
Isotropic splats & 0.725 & 0.499 \\
Sampling-aware splats & 0.742 & 0.510 \\
\midrule
\multicolumn{3}{@{}l}{\emph{\ours{} variants}} \\
w/o learning & 0.357 & 0.316 \\
w/o sampling term & 0.388 & 0.331 \\
\rowcolor{oursbg} \textcolor{oursblue}{\textbf{\ours}} & \textbf{0.275} & \textbf{0.276} \\
\bottomrule
\end{tabularx}
\end{table}

\noindent\textbf{Learned attributes.} In \cref{tab:ablation}, we compare \ours{} w/o learning, which uses fixed attribute rules, with the full model. Learning Gaussian attributes reduces CD from 0.357 to 0.275\,m on PAI (\resultPAIGtwoGain\%) and from 0.316 to 0.276\,m on Waymo (\resultWaymoGtwoGain\%), improving all 12 clips. Both variants use the same measurement anchors, consolidation, and post-processing filter. This demonstrates the importance of the predictor training.

\noindent\textbf{Sampling-aware Gaussian scales.} In \cref{tab:ablation}, replacing Isotropic splats with Sampling-aware splats slightly increases CD from 0.725 to 0.742\,m on PAI and from 0.499 to 0.510\,m on Waymo. This is because expanding spatial support alone still leaves redundant and isolated predictions unresolved; its benefit is realized within the full pipeline with consolidation and filtering. We retrain and evaluate \ours{} without the sampling terms $b_{a,i}$ in \cref{eq:sampling-scale} (\emph{w/o sampling term}), retaining consolidation and the post-processing filter. CD rises from 0.275 to 0.388\,m on PAI and from 0.276 to 0.331\,m on Waymo. This comparison shows that sampling-aware Gaussian scales can help improve coverage and reduce reconstruction error.

\begin{table}[!htbp]\centering
\caption{\textbf{Multi-frame consolidation ablations.} All variants reuse identical protocols. CD is in meters; GS counts include static and dynamic primitives. No fusion retains source selection and skips merging. Bold marks the best result within each dataset.}\label{tab:consolidation}
\fontsize{9}{10.4}\selectfont
\setlength{\tabcolsep}{3pt}
\renewcommand{\arraystretch}{1.12}
\begin{tabularx}{\columnwidth}{@{}lRRR@{}}
\toprule
Variant & CD$\downarrow$ & F@5$\uparrow$ & GS (M)$\downarrow$ \\
\midrule
\multicolumn{4}{@{}l}{\emph{PAI}} \\
Registered concatenation & 0.372 & 0.322 & 52.694 \\
Covariance averaging & 0.433 & 0.291 & \textbf{6.704} \\
\rowcolor{oursbg} \textcolor{oursblue}{\textbf{Moment matching}} & \textbf{0.275} & \textbf{0.495} & \textbf{6.704} \\
\midrule
\multicolumn{4}{@{}l}{\emph{Waymo}} \\
Registered concatenation & 0.419 & 0.272 & 13.981 \\
No fusion & \textbf{0.275} & 0.402 & 7.298 \\
Covariance averaging & 0.412 & 0.290 & \textbf{0.780} \\
\rowcolor{oursbg} \textcolor{oursblue}{\textbf{Moment matching}} & 0.276 & \textbf{0.466} & \textbf{0.780} \\
\bottomrule
\end{tabularx}
\end{table}

\noindent\textbf{Multi-frame consolidation.} In \cref{tab:consolidation}, we compare the full pipeline (\emph{Moment matching}) with \emph{Registered concatenation}, which simply registers and concatenates per-frame Gaussians. Consolidation lowers CD and raises F@5 while reducing Gaussian counts by $7.9\times$ on PAI and $17.9\times$ on Waymo. The Waymo \emph{No fusion} variant retains source selection but skips Gaussian merging. Adding fusion reduces the count from 7.298M to 0.780M and improves F@5 from 0.402 to 0.466, while CD remains nearly unchanged (0.275 to 0.276\,m). 

\noindent\textbf{Moment matching.} In \cref{tab:consolidation}, we also compare covariance averaging and moment matching with all other consolidation steps fixed. Both produce identical merged centers, but moment matching also preserves center spread (\cref{fig:consolidation}(c,d)). Using moment matching raises F@5 from 0.291 to 0.495 on PAI and from 0.290 to 0.466 on Waymo, improving every clip with identical Gaussian counts. 

\noindent\textbf{Limitations.} \ours{} still trails the best per-scene optimization methods in reconstruction metrics, despite faster reconstruction and stronger downstream detection on held-out and laterally shifted views. \ours{} relies on accurate sensor poses and dynamic-object annotations, as errors can misalign Gaussians during multi-frame consolidation and increase reconstruction error. The pipeline is not integrated with camera-based 3DGS, so joint camera--LiDAR simulation requires deploying separate models and generating two sets of assets; unifying these models and assets is a direction for future work.
\endgroup

\section{Conclusion}\label{sec:conclusion}
\ours{} combines measurement-anchored Gaussian prediction, sampling-aware Gaussian scales, and multi-frame consolidation to reconstruct reusable LiDAR assets in minutes without per-scene optimization. On PAI and Waymo, reconstruction is approximately $46\times$ and $70\times$ faster than NuRec, respectively, with competitive downstream detection on Waymo. On held-out frames and at a 3\,m lateral offset, it achieves the highest pooled vehicle and pedestrian AP among the compared reconstructions. Lower reconstruction costs can help automotive developers expand simulation libraries and scale closed-loop testing, accelerating autonomous driving development. %

{\small\setlength{\baselineskip}{10.5pt}\setlength{\bibsep}{0pt}%
\bibliographystyle{ieeenat_fullname}\bibliography{references}}

\begin{thebibliography}{44}
\providecommand{\natexlab}[1]{#1}
\providecommand{\url}[1]{\texttt{#1}}
\expandafter\ifx\csname urlstyle\endcsname\relax
  \providecommand{\doi}[1]{doi: #1}\else
  \providecommand{\doi}{doi: \begingroup \urlstyle{rm}\Url}\fi

\bibitem[Amini et~al.(2022)Amini, Wang, Gilitschenski, Schwarting, Liu, Han,
  Karaman, and Rus]{vista2}
Alexander Amini, Tsun-Hsuan Wang, Igor Gilitschenski, Wilko Schwarting, Zhijian
  Liu, Song Han, Sertac Karaman, and Daniela Rus.
\newblock {VISTA 2.0}: An open, data-driven simulator for multimodal sensing
  and policy learning for autonomous vehicles.
\newblock In \emph{Proceedings of the IEEE International Conference on Robotics
  and Automation}, pages 2419--2426. IEEE, 2022.

\bibitem[Barron et~al.(2021)Barron, Mildenhall, Tancik, Hedman, Martin-Brualla,
  and Srinivasan]{mipnerf}
Jonathan~T. Barron, Ben Mildenhall, Matthew Tancik, Peter Hedman, Ricardo
  Martin-Brualla, and Pratul~P. Srinivasan.
\newblock {Mip-NeRF}: A multiscale representation for anti-aliasing neural
  radiance fields.
\newblock In \emph{Proceedings of the IEEE/CVF International Conference on
  Computer Vision}, pages 5855--5864, 2021.

\bibitem[Charatan et~al.(2024)Charatan, Li, Tagliasacchi, and
  Sitzmann]{pixelsplat}
David Charatan, Sizhe~Lester Li, Andrea Tagliasacchi, and Vincent Sitzmann.
\newblock {pixelSplat}: {3D} gaussian splats from image pairs for scalable
  generalizable {3D} reconstruction.
\newblock In \emph{Proceedings of the IEEE/CVF Conference on Computer Vision
  and Pattern Recognition}, pages 19457--19467, 2024.

\bibitem[Chen et~al.(2025)Chen, Du, Tang, Xie, Chen, Huo, and Yang]{lidargs}
Qifeng Chen, Sicong Du, Tao Tang, Rengan Xie, Peng Chen, Yuchi Huo, and Sheng
  Yang.
\newblock {LiDAR-GS}: Real-time {LiDAR} re-simulation using gaussian splatting.
\newblock arXiv preprint arXiv:2410.05111v3, 2025.
\newblock Earlier version: arXiv:2410.05111v1, 2024.

\bibitem[Chen et~al.(2026)Chen, Liu, Xie, Tang, Du, Zhao, Huo, and
  Yang]{lidargspp}
Qifeng Chen, Jiarun Liu, Rengan Xie, Tao Tang, Sicong Du, Yiru Zhao, Yuchi Huo,
  and Sheng Yang.
\newblock {LiDAR-GS++}: Improving {LiDAR} gaussian reconstruction via diffusion
  priors.
\newblock In \emph{Proceedings of the AAAI Conference on Artificial
  Intelligence}, pages 2975--2983, 2026.

\bibitem[Chen et~al.(2024)Chen, Xu, Zheng, Zhuang, Pollefeys, Geiger, Cham, and
  Cai]{mvsplat}
Yuedong Chen, Haofei Xu, Chuanxia Zheng, Bohan Zhuang, Marc Pollefeys, Andreas
  Geiger, Tat-Jen Cham, and Jianfei Cai.
\newblock {MVSplat}: Efficient {3D} gaussian splatting from sparse multi-view
  images.
\newblock In \emph{Proceedings of the European Conference on Computer Vision},
  pages 370--386, 2024.

\bibitem[Hess et~al.(2025)Hess, Lindstr{\"o}m, Fatemi, Petersson, and
  Svensson]{splatad}
Georg Hess, Carl Lindstr{\"o}m, Maryam Fatemi, Christoffer Petersson, and
  Lennart Svensson.
\newblock {SplatAD}: Real-time lidar and camera rendering with {3D} gaussian
  splatting for autonomous driving.
\newblock In \emph{Proceedings of the IEEE/CVF Conference on Computer Vision
  and Pattern Recognition}, pages 11982--11992, 2025.

\bibitem[Huang et~al.(2024)Huang, Yu, Chen, Geiger, and Gao]{2dgs}
Binbin Huang, Zehao Yu, Anpei Chen, Andreas Geiger, and Shenghua Gao.
\newblock {2D} gaussian splatting for geometrically accurate radiance fields.
\newblock In \emph{ACM SIGGRAPH Conference Papers}. Association for Computing
  Machinery, 2024.

\bibitem[Huang et~al.(2023)Huang, Gojcic, Wang, Williams, Kasten, Fidler,
  Schindler, and Litany]{nfl}
Shengyu Huang, Zan Gojcic, Zian Wang, Francis Williams, Yoni Kasten, Sanja
  Fidler, Konrad Schindler, and Or Litany.
\newblock Neural {LiDAR} fields for novel view synthesis.
\newblock In \emph{Proceedings of the IEEE/CVF International Conference on
  Computer Vision}, pages 18236--18246, 2023.

\bibitem[Huang et~al.(2026)Huang, Kang, Zhang, Ren, Zhang, and Lai]{lidarevs}
Yiming Huang, Xin Kang, Sipeng Zhang, Hongliang Ren, Weihua Zhang, and Junjie
  Lai.
\newblock {LiDAR-EVS}: Enhance extrapolated view synthesis for {3D} gaussian
  splatting with pseudo-{LiDAR} supervision.
\newblock In \emph{Proceedings of the European Conference on Computer Vision},
  pages 1--18, 2026.

\bibitem[Jiang et~al.(2025)Jiang, Gu, Chen, and Zhang]{gslidar}
Junzhe Jiang, Chun Gu, Yurui Chen, and Li Zhang.
\newblock {GS-LiDAR}: Generating realistic {LiDAR} point clouds with panoramic
  gaussian splatting.
\newblock In \emph{Proceedings of the International Conference on Learning
  Representations}, 2025.

\bibitem[Kerbl et~al.(2023)Kerbl, Kopanas, Leimk{\"u}hler, and
  Drettakis]{kerbl3Dgaussians}
Bernhard Kerbl, Georgios Kopanas, Thomas Leimk{\"u}hler, and George Drettakis.
\newblock {3D} gaussian splatting for real-time radiance field rendering.
\newblock \emph{ACM Transactions on Graphics}, 42\penalty0 (4), 2023.

\bibitem[Kingma and Ba(2015)]{adam}
Diederik~P. Kingma and Jimmy Ba.
\newblock Adam: A method for stochastic optimization.
\newblock In \emph{Proceedings of the International Conference on Learning
  Representations}, 2015.

\bibitem[Knapitsch et~al.(2017)Knapitsch, Park, Zhou, and
  Koltun]{tanksandtemples}
Arno Knapitsch, Jaesik Park, Qian-Yi Zhou, and Vladlen Koltun.
\newblock Tanks and temples: Benchmarking large-scale scene reconstruction.
\newblock \emph{ACM Transactions on Graphics}, 36\penalty0 (4), 2017.

\bibitem[Li et~al.(2023)Li, Ren, and Liu]{pcgen}
Chenqi Li, Yuan Ren, and Bingbing Liu.
\newblock {PCGen}: Point cloud generator for {LiDAR} simulation.
\newblock In \emph{Proceedings of the IEEE International Conference on Robotics
  and Automation}, pages 11676--11682, 2023.

\bibitem[Li et~al.(2026)Li, Li, Gu, Zheng, Yu, and Bai]{lgs}
Hang Li, Jiahe Li, Meiying Gu, Jin Zheng, Lina Yu, and Xiao Bai.
\newblock Learning gaussian structure: Intervention-guided density control for
  feed-forward driving reconstruction.
\newblock arXiv preprint arXiv:2608.11077, 2026.

\bibitem[Lu et~al.(2024)Lu, Yu, Xu, Xiangli, Wang, Lin, and Dai]{scaffoldgs}
Tao Lu, Mulin Yu, Linning Xu, Yuanbo Xiangli, Limin Wang, Dahua Lin, and Bo
  Dai.
\newblock {Scaffold-GS}: Structured {3D} gaussians for view-adaptive rendering.
\newblock In \emph{Proceedings of the IEEE/CVF Conference on Computer Vision
  and Pattern Recognition}, pages 20654--20664, 2024.

\bibitem[Manivasagam et~al.(2020)Manivasagam, Wang, Wong, Zeng, Sazanovich,
  Tan, Yang, Ma, and Urtasun]{lidarsim}
Sivabalan Manivasagam, Shenlong Wang, Kelvin Wong, Wenyuan Zeng, Mikita
  Sazanovich, Shuhan Tan, Bin Yang, Wei-Chiu Ma, and Raquel Urtasun.
\newblock {LiDARsim}: Realistic {LiDAR} simulation by leveraging the real
  world.
\newblock In \emph{Proceedings of the IEEE/CVF Conference on Computer Vision
  and Pattern Recognition}, pages 11167--11176, 2020.

\bibitem[Miao et~al.(2025)Miao, Huang, Bai, Yan, Zhou, Wang, Liu, Geiger, and
  Liao]{evolsplat}
Sheng Miao, Jiaxin Huang, Dongfeng Bai, Xu Yan, Hongyu Zhou, Yue Wang, Bingbing
  Liu, Andreas Geiger, and Yiyi Liao.
\newblock {EVolSplat}: Efficient volume-based gaussian splatting for urban view
  synthesis.
\newblock In \emph{Proceedings of the IEEE/CVF Conference on Computer Vision
  and Pattern Recognition}, pages 11286--11296, 2025.

\bibitem[{NVIDIA}(2025)]{cosmoslidargen}
{NVIDIA}.
\newblock {Cosmos-Transfer-LidarGen}: A diffusion model for {LiDAR} generation.
\newblock Hugging Face model card, 2025.
\newblock Released October 23, 2025; accessed November 11, 2026.

\bibitem[{NVIDIA}(2026)]{nurec}
{NVIDIA}.
\newblock Reconstruct an {AV} scene with {NuRec}.
\newblock NVIDIA Omniverse NuRec documentation, 2026.
\newblock Accessed November 4, 2026.

\bibitem[Ost et~al.(2021)Ost, Mannan, Thuerey, Knodt, and Heide]{nsg}
Julian Ost, Fahim Mannan, Nils Thuerey, Julian Knodt, and Felix Heide.
\newblock Neural scene graphs for dynamic scenes.
\newblock In \emph{Proceedings of the IEEE/CVF Conference on Computer Vision
  and Pattern Recognition}, pages 2856--2865, 2021.

\bibitem[Ranftl et~al.(2021)Ranftl, Bochkovskiy, and Koltun]{dpt}
Ren{\'e} Ranftl, Alexey Bochkovskiy, and Vladlen Koltun.
\newblock Vision transformers for dense prediction.
\newblock In \emph{Proceedings of the IEEE/CVF International Conference on
  Computer Vision}, pages 12179--12188, 2021.

\bibitem[Runnalls(2007)]{runnalls}
Andrew~R. Runnalls.
\newblock {Kullback-Leibler} approach to {Gaussian} mixture reduction.
\newblock \emph{IEEE Transactions on Aerospace and Electronic Systems},
  43\penalty0 (3):\penalty0 989--999, 2007.

\bibitem[Sun et~al.(2020)Sun, Kretzschmar, Dotiwalla, Chouard, Patnaik, Tsui,
  Guo, Zhou, Chai, Caine, Vasudevan, Han, Ngiam, Zhao, Timofeev, Ettinger,
  Krivokon, Gao, Joshi, Zhang, Shlens, Chen, and Anguelov]{waymo}
Pei Sun, Henrik Kretzschmar, Xerxes Dotiwalla, Aurelien Chouard, Vijaysai
  Patnaik, Paul Tsui, James Guo, Yin Zhou, Yuning Chai, Benjamin Caine, Vijay
  Vasudevan, Wei Han, Jiquan Ngiam, Hang Zhao, Aleksei Timofeev, Scott
  Ettinger, Maxim Krivokon, Amy Gao, Aditya Joshi, Yu Zhang, Jonathon Shlens,
  Zhifeng Chen, and Dragomir Anguelov.
\newblock Scalability in perception for autonomous driving: Waymo open dataset.
\newblock In \emph{Proceedings of the IEEE/CVF Conference on Computer Vision
  and Pattern Recognition}, pages 2446--2454, 2020.

\bibitem[Tao et~al.(2024{\natexlab{a}})Tao, Gao, Wang, Lao, Chen, Zhao, Hao,
  Liang, Salzmann, and Yu]{lidarnerf}
Tang Tao, Longfei Gao, Guangrun Wang, Yixing Lao, Peng Chen, Hengshuang Zhao,
  Dayang Hao, Xiaodan Liang, Mathieu Salzmann, and Kaicheng Yu.
\newblock {LiDAR-NeRF}: Novel {LiDAR} view synthesis via neural radiance
  fields.
\newblock In \emph{Proceedings of the ACM International Conference on
  Multimedia}, pages 390--398, 2024{\natexlab{a}}.

\bibitem[Tao et~al.(2024{\natexlab{b}})Tao, Wang, Lao, Chen, Liu, Lin, Yu, and
  Liang]{alignmif}
Tang Tao, Guangrun Wang, Yixing Lao, Peng Chen, Jie Liu, Liang Lin, Kaicheng
  Yu, and Xiaodan Liang.
\newblock {AlignMiF}: Geometry-aligned multimodal implicit field for
  {LiDAR}-camera joint synthesis.
\newblock In \emph{Proceedings of the IEEE/CVF Conference on Computer Vision
  and Pattern Recognition}, pages 21230--21240, 2024{\natexlab{b}}.

\bibitem[Tian et~al.(2025)Tian, Tan, Xie, and Ma]{drivingforward}
Qijian Tian, Xin Tan, Yuan Xie, and Lizhuang Ma.
\newblock {DrivingForward}: Feed-forward {3D} gaussian splatting for driving
  scene reconstruction from flexible surround-view input.
\newblock In \emph{Proceedings of the AAAI Conference on Artificial
  Intelligence}, pages 7374--7382, 2025.

\bibitem[Tonderski et~al.(2024)Tonderski, Lindstr{\"o}m, Hess, Ljungbergh,
  Svensson, and Petersson]{neurad}
Adam Tonderski, Carl Lindstr{\"o}m, Georg Hess, William Ljungbergh, Lennart
  Svensson, and Christoffer Petersson.
\newblock {NeuRAD}: Neural rendering for autonomous driving.
\newblock In \emph{Proceedings of the IEEE/CVF Conference on Computer Vision
  and Pattern Recognition}, pages 14895--14904, 2024.

\bibitem[Vaswani et~al.(2017)Vaswani, Shazeer, Parmar, Uszkoreit, Jones, Gomez,
  Kaiser, and Polosukhin]{attention}
Ashish Vaswani, Noam Shazeer, Niki Parmar, Jakob Uszkoreit, Llion Jones,
  Aidan~N. Gomez, {\L}ukasz Kaiser, and Illia Polosukhin.
\newblock Attention is all you need.
\newblock In \emph{Advances in Neural Information Processing Systems}, 2017.

\bibitem[Wang et~al.(2026)Wang, Sun, Bai, Casser, Peng, Zhu, Shih, Masotto, Su,
  Parvate, Ge, Bieske, Anguelov, Tan, and Jiang]{sensor2sensor}
Jiahao Wang, Bo Sun, Yijing Bai, Vincent Casser, Songyou Peng, Zehao Zhu,
  Meng-Li Shih, Xander Masotto, Shih-Yang Su, Kanaad Parvate, Tiancheng Ge,
  Linn Bieske, Dragomir Anguelov, Mingxing Tan, and Chiyu~Max Jiang.
\newblock {Sensor2Sensor}: Cross-embodiment sensor conversion for autonomous
  driving.
\newblock In \emph{Proceedings of the IEEE/CVF Conference on Computer Vision
  and Pattern Recognition}, 2026.

\bibitem[Wu et~al.(2024)Wu, Zuo, Leutenegger, Litany, Schindler, and
  Huang]{dynfl}
Hanfeng Wu, Xingxing Zuo, Stefan Leutenegger, Or Litany, Konrad Schindler, and
  Shengyu Huang.
\newblock Dynamic {LiDAR} re-simulation using compositional neural fields.
\newblock In \emph{Proceedings of the IEEE/CVF Conference on Computer Vision
  and Pattern Recognition}, pages 19988--19998, 2024.

\bibitem[Wu et~al.(2025)Wu, Esturo, Mirzaei, Mo{\"e}nne-Loccoz, and
  Gojcic]{3dgut}
Qi Wu, Janick~Martinez Esturo, Ashkan Mirzaei, Nicolas Mo{\"e}nne-Loccoz, and
  Zan Gojcic.
\newblock {3DGUT}: Enabling distorted cameras and secondary rays in gaussian
  splatting.
\newblock In \emph{Proceedings of the IEEE/CVF Conference on Computer Vision
  and Pattern Recognition}, pages 26036--26046, 2025.

\bibitem[Xu et~al.(2025)Xu, Peng, Wang, Blum, Barath, Geiger, and
  Pollefeys]{depthsplat}
Haofei Xu, Songyou Peng, Fangjinhua Wang, Hermann Blum, Daniel Barath, Andreas
  Geiger, and Marc Pollefeys.
\newblock {DepthSplat}: Connecting gaussian splatting and depth.
\newblock In \emph{Proceedings of the IEEE/CVF Conference on Computer Vision
  and Pattern Recognition}, pages 16453--16463, 2025.

\bibitem[Yan et~al.(2024)Yan, Lin, Zhou, Wang, Sun, Zhan, Lang, Zhou, and
  Peng]{streetgaussians}
Yunzhi Yan, Haotong Lin, Chenxu Zhou, Weijie Wang, Haiyang Sun, Kun Zhan,
  Xianpeng Lang, Xiaowei Zhou, and Sida Peng.
\newblock Street gaussians: Modeling dynamic urban scenes with gaussian
  splatting.
\newblock In \emph{Proceedings of the European Conference on Computer Vision},
  pages 156--173, 2024.

\bibitem[Yang et~al.(2025)Yang, Huang, Ivanovic, Chen, Wang, Li, You, Sharma,
  Igl, Karkus, Xu, Wang, and Pavone]{storm}
Jiawei Yang, Jiahui Huang, Boris Ivanovic, Yuxiao Chen, Yan Wang, Boyi Li,
  Yurong You, Apoorva Sharma, Maximilian Igl, Peter Karkus, Danfei Xu, Yue
  Wang, and Marco Pavone.
\newblock {STORM}: Spatio-temporal reconstruction model for large-scale outdoor
  scenes.
\newblock In \emph{Proceedings of the International Conference on Learning
  Representations}, 2025.

\bibitem[Yang et~al.(2023)Yang, Chen, Wang, Manivasagam, Ma, Yang, and
  Urtasun]{unisim}
Ze Yang, Yun Chen, Jingkang Wang, Sivabalan Manivasagam, Wei-Chiu Ma,
  Anqi~Joyce Yang, and Raquel Urtasun.
\newblock {UniSim}: A neural closed-loop sensor simulator.
\newblock In \emph{Proceedings of the IEEE/CVF Conference on Computer Vision
  and Pattern Recognition}, pages 1389--1399, 2023.

\bibitem[Yin et~al.(2021)Yin, Zhou, and Kr{\"a}henb{\"u}hl]{centerpoint}
Tianwei Yin, Xingyi Zhou, and Philipp Kr{\"a}henb{\"u}hl.
\newblock Center-based {3D} object detection and tracking.
\newblock In \emph{Proceedings of the IEEE/CVF Conference on Computer Vision
  and Pattern Recognition}, pages 11784--11793, 2021.

\bibitem[Yu et~al.(2024)Yu, Chen, Huang, Sattler, and Geiger]{mipsplatting}
Zehao Yu, Anpei Chen, Binbin Huang, Torsten Sattler, and Andreas Geiger.
\newblock Mip-splatting: Alias-free {3D} gaussian splatting.
\newblock In \emph{Proceedings of the IEEE/CVF Conference on Computer Vision
  and Pattern Recognition}, pages 19447--19456, 2024.

\bibitem[Zheng et~al.(2024)Zheng, Lu, Xue, Chen, and Jiang]{lidar4d}
Zehan Zheng, Fan Lu, Weiyi Xue, Guang Chen, and Changjun Jiang.
\newblock {LiDAR4D}: Dynamic neural fields for novel space-time view {LiDAR}
  synthesis.
\newblock In \emph{Proceedings of the IEEE/CVF Conference on Computer Vision
  and Pattern Recognition}, pages 5145--5154, 2024.

\bibitem[Zhou et~al.(2025)Zhou, Fu, Peng, Yan, Zhang, Chen, Xia, and
  Zhou]{lidarrt}
Chenxu Zhou, Lvchang Fu, Sida Peng, Yunzhi Yan, Zhanhua Zhang, Yong Chen,
  Jiazhi Xia, and Xiaowei Zhou.
\newblock {LiDAR-RT}: Gaussian-based ray tracing for dynamic {LiDAR}
  re-simulation.
\newblock In \emph{Proceedings of the IEEE/CVF Conference on Computer Vision
  and Pattern Recognition}, pages 1538--1548, 2025.

\bibitem[Zhu et~al.(2026)Zhu, Li, Zhou, Li, and Liu]{nurecscaling}
Tyler Zhu, Phoebe Li, Mofan Zhou, Yimeng Li, and Chen Liu.
\newblock Scaling neural reconstruction for closed-loop autonomous vehicle
  simulation.
\newblock Best-practices report, NVIDIA and Li Auto, 2026.
\newblock Accessed November 4, 2026.

\bibitem[Zwicker et~al.(2002)Zwicker, Pfister, van Baar, and Gross]{ewa}
Matthias Zwicker, Hanspeter Pfister, Jeroen van Baar, and Markus Gross.
\newblock {EWA} splatting.
\newblock \emph{IEEE Transactions on Visualization and Computer Graphics},
  8\penalty0 (3):\penalty0 223--238, 2002.

\bibitem[Zyrianov et~al.(2022)Zyrianov, Zhu, and Wang]{lidargen}
Vlas Zyrianov, Xiyue Zhu, and Shenlong Wang.
\newblock Learning to generate realistic {LiDAR} point clouds.
\newblock In \emph{Proceedings of the European Conference on Computer Vision},
  2022.

\end{thebibliography}

\clearpage
\appendix
\setcounter{table}{0}
\setcounter{figure}{0}
\setcounter{equation}{0}
\renewcommand{\thetable}{S\arabic{table}}
\renewcommand{\thefigure}{S\arabic{figure}}
\renewcommand{\theequation}{S\arabic{equation}}
\renewcommand{\theHtable}{supp.\arabic{table}}
\renewcommand{\theHfigure}{supp.\arabic{figure}}
\renewcommand{\theHequation}{supp.\arabic{equation}}
\maketitlesupplementary
\raggedbottom
\clubpenalty=10000
\brokenpenalty=10000
\section{Network architecture}\label{sec:supp-architecture}
\textbf{LiDAR input and encoder.}
Each training window contains $8$--$18$ LiDAR scans, their sensor poses, and timestamps. We store each scan as a two-channel map containing range and intensity. Before feeding the map to the encoder, we clip ranges to $[0,200]$\,m and map them linearly to $[-1,1]$. We clip intensities to $[0,1]$. The LiDAR encoder splits each map into patches of $16\times30$ pixels and maps each patch to a 192-dimensional token. \Cref{tab:supp-architecture} lists the architecture settings.

\textbf{Pose and time encoder.}
Each frame also provides an eight-dimensional vector, including three translation coordinates, a four-dimensional rotation quaternion, and one timestamp relative to the first input frame, in seconds. An MLP maps this vector through a 96-dimensional hidden layer to a 192-dimensional token. Four self-attention blocks, each with 16 heads, then exchange information between the pose and time tokens of different input frames. The output is one pose and time token per frame.

\textbf{Transformer.}
The main transformer has eight blocks, with six attention heads and 192-dimensional tokens. Blocks 1 and 2 process each frame separately. We prepend each frame’s pose and time token to its LiDAR patch tokens once, at the input to block 3. These tokens are then processed together throughout blocks 3–8. Blocks 3, 5, and 7 apply attention within each frame; blocks 4, 6, and 8 apply attention across all input frames. 
The transformer combines information within and across LiDAR scans to produce features for Gaussian attribute prediction.

\textbf{DPT head.}
We extract patch features at blocks 3, 5, 7, and 8 to form four inputs to DPT~\cite{dpt}. Each input concatenates two 192-channel feature maps along the channel dimension, giving 384 channels. At blocks 3, 5, and 7, we concatenate that block's output with its LayerNorm-normalized version. At block 8, we concatenate the output of block 7 with the normalized output of block 8. DPT projects these four inputs to the channel widths listed in \cref{tab:supp-architecture} and resizes them to different spatial resolutions. It then fuses and upsamples them to the original range-map resolution. It predicts position and intensity residuals, scales, opacities, rotation residuals, and the coefficients used for sampling-aware Gaussian scales. The sampling coefficient is $k=0.3(1+0.5\tanh\kappa)$, where $\kappa$ is a network output. Rotation residuals are initialized to the identity rotation.

\begin{table}[!t]
\centering
\caption{\textbf{Network architecture.} Input map shapes list beam rows, azimuth columns, and input channels.}
\label{tab:supp-architecture}
\small
\setlength{\tabcolsep}{4pt}
\renewcommand{\arraystretch}{1.0}
\begin{tabularx}{\linewidth}{@{}>{\raggedright\arraybackslash}Xr@{}}
\toprule
\textbf{Parameter} & \textbf{Value} \\
\midrule
\rowcolor{oursbg}[0pt][0pt]
\multicolumn{2}{@{}l@{}}{\color{oursblue!75!black}\bfseries Inputs}\\
PAI input map shape & $128\times3600\times2$ \\
Waymo input map shape & $64\times2640\times2$ \\
Translation dimension & 3 \\
Quaternion dimension & 4 \\
Relative-time dimension & 1 \\
\addlinespace[1pt]
\rowcolor{oursbg}[0pt][0pt]
\multicolumn{2}{@{}l@{}}{\color{oursblue!75!black}\bfseries LiDAR encoder}\\
Patch size (rows, columns) & $16\times30$ \\
Patch token dimension & 192 \\
\addlinespace[1pt]
\rowcolor{oursbg}[0pt][0pt]
\multicolumn{2}{@{}l@{}}{\color{oursblue!75!black}\bfseries Pose and time encoder}\\
MLP hidden dimension & 96 \\
Cross-frame attention blocks & 4 \\
Attention heads & 16 \\
Pose and time token dimension & 192 \\
\addlinespace[1pt]
\rowcolor{oursbg}[0pt][0pt]
\multicolumn{2}{@{}l@{}}{\color{oursblue!75!black}\bfseries  Transformer}\\
Attention blocks & 8 \\
Token dimension & 192 \\
Attention heads & 6 \\
MLP expansion ratio & 4 \\
Rotary position encoding base & 100 \\
\addlinespace[1pt]
\rowcolor{oursbg}[0pt][0pt]
\multicolumn{2}{@{}l@{}}{\color{oursblue!75!black}\bfseries DPT head}\\
Source Transformer blocks & $(3,5,7,8)$ \\
Input channels per level & 384 \\
Projected channels per level & $(96,192,384,384)$ \\
Fused feature channels & 128 \\
\bottomrule
\end{tabularx}
\end{table}

\section{Training and inference}\label{sec:implementation}
Algorithm~1 outlines the training and inference workflows described below.

\begin{figure*}[!t]
\small
\setlength{\parindent}{0pt}
\renewcommand{\arraystretch}{1.05}
\setlength{\fboxsep}{3pt}
\hrule height 0.8pt
\vspace{5pt}
\textbf{Algorithm 1}\quad Training and inference
\par\vspace{5pt}\hrule\vspace{5pt}
\begin{minipage}[t]{0.48\textwidth}
\colorbox{oursbg}{\parbox{\dimexpr\linewidth-2\fboxsep\relax}{\color{oursblue!75!black}\textbf{(a) Training}}}\par\vspace{4pt}
\textbf{Input:} Training clips with sensor calibration, poses, timestamps, and actor tracks.\par
\textbf{Output:} A trained Gaussian predictor.\par\vspace{4pt}
\textit{Repeat for each training batch:}\par\vspace{3pt}
\begin{tabular}{@{}>{\raggedleft\arraybackslash}p{10pt}@{\hspace{5pt}}>{\raggedright\arraybackslash}p{\dimexpr\linewidth-15pt\relax}@{}}
1 & \textbf{Sample} input windows and separate supervision scans.\\
2 & \textbf{Predict} Gaussians anchored at valid input returns.\\
3 & \textbf{Render} at the supervision poses and times, using actor tracks to move dynamic Gaussians.\\
4 & \textbf{Compute} range, intensity, and opacity losses on valid supervision rays.\\
5 & \textbf{Update} the predictor with Adam by backpropagating the combined loss.\\
\end{tabular}
\end{minipage}\hfill
\begin{minipage}[t]{0.48\textwidth}
\colorbox{oursbg}{\parbox{\dimexpr\linewidth-2\fboxsep\relax}{\color{oursblue!75!black}\textbf{(b) Inference}}}\par\vspace{4pt}
\textbf{Input:} A new clip, the trained predictor, query poses and times.\par
\textbf{Output:} A Gaussian scene and rendered point clouds.\par\vspace{4pt}
\textit{Keep the predictor weights fixed:}\par\vspace{3pt}
\begin{tabular}{@{}>{\raggedleft\arraybackslash}p{10pt}@{\hspace{5pt}}>{\raggedright\arraybackslash}p{\dimexpr\linewidth-15pt\relax}@{}}
1 & \textbf{Predict} Gaussians for each input window.\\
2 & \textbf{Align} static Gaussians using sensor poses and associate dynamic Gaussians with actor tracks.\\
3 & \textbf{Consolidate} static Gaussians: select sources in each grid cell, merge by moment matching, and remove isolated centers.\\
4 & \textbf{Assemble} static and dynamic Gaussians into a scene.\\
5 & \textbf{Render} at query poses and times, using actor tracks to move dynamic Gaussians.\\
6 & \textbf{Filter} returns with low accumulated opacity or high depth dispersion.\\
\end{tabular}
\end{minipage}
\par\vspace{6pt}\hrule height 0.8pt
\end{figure*}

\subsection{Training}\label{sec:supp-training}
In Algorithm~1(a), each training example contains input scans for Gaussian prediction and separate supervision scans for rendering. These two sets do not overlap. The predictor instantiates Gaussians at valid input returns. We render range, intensity, and accumulated opacity at the supervision poses and timestamps, using actor tracks to move dynamic Gaussians, and minimize
\begin{equation}
\mathcal L=\mathcal L_{\mathrm{range}}+\mathcal L_{\mathrm{intensity}}+0.1\mathcal L_{\mathrm{opacity}}.
\end{equation}
The three terms are the mean absolute range error, mean squared intensity error, and mean absolute error between accumulated opacity and one. All three are evaluated on rays with recorded returns in the supervision scans. We do not train a separate ray-drop classifier.

Adam jointly updates both input encoders, the main transformer, and the DPT head by backpropagating this loss. \Cref{tab:supp-hyperparameters} lists the settings. We train separate PAI and Waymo predictors for two epochs, using a global batch of 32 input windows and BF16 mixed precision on H20 GPUs. The learning rate warms up from $10^{-6}$ to $10^{-4}$ over 100 steps, then decays to $3.33\times10^{-6}$ with a cosine schedule.

\begin{table}[!t]
\centering
\caption{\textbf{Prediction and training settings.} Batch size counts input windows across all GPUs.}
\label{tab:supp-hyperparameters}
\small
\setlength{\tabcolsep}{4pt}
\renewcommand{\arraystretch}{1.0}
\begin{tabularx}{\linewidth}{@{}>{\raggedright\arraybackslash}Xr@{}}
\toprule
\textbf{Parameter} & \textbf{Value} \\
\midrule
\rowcolor{oursbg}[0pt][0pt]
\multicolumn{2}{@{}l@{}}{\color{oursblue!75!black}\bfseries Gaussian prediction}\\
Max. position residual per axis (m) & 0.5 \\
Maximum intensity residual & 0.1 \\
\addlinespace[1pt]
\rowcolor{oursbg}[0pt][0pt]
\multicolumn{2}{@{}l@{}}{\color{oursblue!75!black}\bfseries Sampling-aware Gaussian scales}\\
Coefficient base / residual factor & $0.3\;/\;0.5$ \\
Thickness floor $b_{\mathrm{rad}}$ (m) & $0.02$ \\
\addlinespace[1pt]
\rowcolor{oursbg}[0pt][0pt]
\multicolumn{2}{@{}l@{}}{\color{oursblue!75!black}\bfseries Training setup}\\
Training epochs & 2 \\
Input windows per global batch & 32 \\
Windows per clip per frame spacing & 10 \\
Input scans per window & $8$--$18$ \\
Time between input scans (s) & $\{0.25,0.5\}$ \\
Supervision scans per window & $8$ \\
\addlinespace[1pt]
\rowcolor{oursbg}[0pt][0pt]
\multicolumn{2}{@{}l@{}}{\color{oursblue!75!black}\bfseries Optimization}\\
Initial / peak learning rate & $10^{-6}\;/\;10^{-4}$ \\
Final learning rate & $3.33\times10^{-6}$ \\
Learning-rate warmup steps & 100 \\
Adam $(\beta_1,\beta_2,\epsilon)$ & $(0.9,0.99,10^{-15})$ \\
\addlinespace[1pt]
\rowcolor{oursbg}[0pt][0pt]
\multicolumn{2}{@{}l@{}}{\color{oursblue!75!black}\bfseries Loss weights}\\
Range / intensity / opacity & $1\;/\;1\;/\;0.1$ \\
\bottomrule
\end{tabularx}
\end{table}

\subsection{Inference}\label{sec:supp-inference}
We predict Gaussians for each input window with fixed predictor weights (Algorithm~1(b)). Given sensor poses, we align static centers and covariances in the scene coordinate system. Given actor tracks, we associate dynamic Gaussians for moving objects.

We then consolidate the static Gaussians in each $0.05$\,m cubic grid cell. A Gaussian's source distance is its center's distance to the nearest sensor position in its input window. Within each cell, we retain candidates whose source distance is at most $1$\,m greater than the cell's minimum, and discard cells observed in only one input window. We merge the remaining Gaussians with equal weights using moment matching. The fused covariance includes both the input covariances and the spread of their centers. After fusion, we remove isolated centers by filtering out static centers that has fewer than 15 static centers within a $0.5$\,m radius. Centers farther than $40$\,m from all input sensors are retained.

We combine the consolidated static Gaussians and tracked dynamic Gaussians into a scene, then render at the query poses and times. Actor tracks move dynamic Gaussians as described in \cref{sec:supp-dynamics}. We remove rendered returns with accumulated opacity below $0.1$ or high weighted depth standard deviation. The latter indicates that the contributing Gaussians lie at different depths. Its threshold is the 95th percentile of these deviations in the query scan, clamped to $[0.3,1.5]$\,m. This step filters the rendered point cloud without changing the stored Gaussian scene.

\section{Dataset splits and evaluation frames}\label{sec:supp-splits}
\textbf{PAI.}
This non-public driving Physical AI dataset uses a 128-beam LiDAR. We train on 723 clips and randomly select six evaluation clips from a disjoint test set. Each evaluation clip supplies 87 or 89 input scans for reconstruction. We hold out frames $5,15,\ldots,95$, giving ten queries per clip and 60 queries in total. \Cref{tab:supp-clips} lists the evaluation clip IDs.

\textbf{Waymo.}
We follow the evaluation segment selection of LiDAR-GS~\cite{lidargs} (v1). Its eight evaluation segments are reserved from Waymo's official 798-segment training split. We sort the remaining 790 segment IDs lexicographically and assign entries $40,80,\ldots,760$ to validation. This gives 19 validation segments and 771 training segments. Six of the eight reserved segments are used for evaluation. One excluded segment (ID prefix \texttt{113792}) has no annotated detection objects in the 54-frame evaluation window. In the other (\texttt{117240}), an actor track ends before three of the four held-out queries. At these queries, the other methods skip rendering the actor outside its valid time window, whereas the evaluated SplatAD~\cite{splatad} and LiDAR-EVS~\cite{lidarevs} implementations would continue rendering it at its last observed pose. This difference in handling track validity would affect evaluation fairness, so we exclude the segment for all methods. Reconstruction and detection use the same six segments. All reserved segments are excluded from predictor training and validation. For each evaluated segment, we use frames 0--53, hold out frames $11,22,32,43$, and reconstruct from the remaining 50 scans. This gives 24 held-out reconstruction queries across six segments.

\begin{table}[!ht]\centering
\caption{\textbf{Evaluation clip IDs.} Six clips are evaluated in each dataset. None is used to train the Gaussian predictor.}
\label{tab:supp-clips}
\small\setlength{\tabcolsep}{2pt}
\begin{tabularx}{\columnwidth}{@{}X@{}}
\toprule
\rowcolor{oursbg}[0pt][0pt] \textcolor{oursblue!75!black}{\textbf{PAI: 128-beam LiDAR}}\\
\texttt{92af11b8-8847-4909-8ef1-3f89f18f4e81}\\
\texttt{48ef7ac2-d61e-4a98-ad95-22af300d96cd}\\
\texttt{8f619ae7-7074-409b-9d05-31f811fd6f86}\\
\texttt{a37339b6-efdf-4b9a-9f0e-f25a9b9631df}\\
\texttt{fb319f97-c892-45e9-a617-4491c6c750bc}\\
\texttt{ff1b156b-2528-44ee-a572-7c471ed83948}\\
\midrule
\rowcolor{oursbg}[0pt][0pt] \textcolor{oursblue!75!black}{\textbf{Waymo: 64-beam TOP LiDAR}}\\
\texttt{10072140764565668044\_4060\_000\_4080\_000}\\
\texttt{10500357041547037089\_1474\_800\_1494\_800}\\
\texttt{10676267326664322837\_311\_180\_331\_180}\\
\texttt{1083056852838271990\_4080\_000\_4100\_000}\\
\texttt{13271285919570645382\_5320\_000\_5340\_000}\\
\texttt{17761959194352517553\_5448\_420\_5468\_420}\\
\bottomrule
\end{tabularx}
\end{table}

\textbf{Detection frames.}
Detection is evaluated at timestamps supported by all compared methods. We use frames 1--52 in five Waymo segments and frames 0--53 in segment (ID prefix \texttt{17761959}). The resulting 314 frames comprise 290 reconstruction-input frames and 24 held-out frames. We report these two groups separately. The shifted-view evaluation uses the same 314 timestamps, moving the sensor $3$\,m to the vehicle's left while keeping its heading and the actor states fixed.

\section{Dynamic object processing}\label{sec:supp-dynamics}
\textbf{Assigning Gaussians to actors.}
We use the annotated 3D box tracks to identify moving objects. A track is treated as dynamic if its annotation marks it as moving. At each return's timestamp, we test whether the predicted Gaussian center lies inside a dynamic actor's padded box. A center inside such a box is assigned to that actor. Unassigned centers remain in the static scene.

\textbf{Moving Gaussians to a query time.}
For each assigned Gaussian, we use the actor track to transform its center from the source timestamp to the previous and next keyframe timestamps. Together with the source center, these positions define three motion keyframes. To render at an intermediate time, we linearly interpolate the center between the two surrounding keyframes. The Gaussian's scale, orientation, and intensity remain fixed during this interpolation. Its opacity fades near either end of its valid time interval.

\textbf{Assembling the dynamic scene.}
Dynamic Gaussians retain their track-based motion when predictions from multiple windows are assembled. They bypass static source selection, moment-matching fusion, and neighborhood filtering. The final scene therefore contains a consolidated static component and dynamic components whose centers vary with query time.

\section{Detector and evaluation settings}\label{sec:supp-detection}
\begin{table*}[!t]\centering
\caption{\textbf{Per-scene Waymo detection.} Vehicle and pedestrian LEVEL\_1 AP ($\uparrow$) for NuRec, LiDAR-GS, and \ours{}. Scene ID prefixes match \cref{tab:supp-clips}. GT counts are annotated box instances summed over frames. Pooled AP is computed over all scenes, not averaged over scene scores. Bold marks the best score among the three methods; a dash denotes no GT instances.}\label{tab:supp-scene-detection}
\fontsize{9}{10.4}\selectfont
\setlength{\tabcolsep}{3pt}\renewcommand{\arraystretch}{1.1}
\begin{tabularx}{\textwidth}{@{}l*{9}{R}@{}}
\toprule
 & & \multicolumn{2}{c}{GT instances} & \multicolumn{2}{c}{NuRec} & \multicolumn{2}{c}{LiDAR-GS} & \multicolumn{2}{c}{\textcolor{oursblue}{\textbf{\ours{}}}} \\
\cmidrule(lr){3-4}\cmidrule(lr){5-6}\cmidrule(lr){7-8}\cmidrule(lr){9-10}
Scene ID & Frames & Vehicle & Pedestrian & Vehicle & Pedestrian & Vehicle & Pedestrian & Vehicle & Pedestrian \\
\midrule
\rowcolor{oursbg}[0pt][0pt] \multicolumn{10}{@{}l@{}}{\textcolor{oursblue!75!black}{\textbf{Held-out recorded-trajectory frames}}}\\
\texttt{10072140} & 4 & 168 & 23 & 0.596 & \textbf{0.321} & 0.626 & 0.154 & \textbf{0.707} & 0.298 \\
\texttt{10500357} & 4 & 170 & 9 & 0.398 & 0.045 & 0.408 & 0.077 & \textbf{0.451} & \textbf{0.080} \\
\texttt{10676267} & 4 & 62 & 3 & \textbf{0.181} & 0.005 & 0.156 & 0.000 & 0.166 & \textbf{0.095} \\
\texttt{10830568} & 4 & 110 & 7 & \textbf{0.651} & \textbf{0.714} & 0.440 & 0.510 & 0.518 & \textbf{0.714} \\
\texttt{13271285} & 4 & 131 & 34 & 0.363 & 0.240 & 0.476 & 0.225 & \textbf{0.671} & \textbf{0.482} \\
\texttt{17761959} & 4 & 30 & 0 & \textbf{0.334} & --- & 0.301 & --- & 0.251 & --- \\
\cmidrule(lr){1-10}
\textbf{Pooled} & 24 & 671 & 76 & 0.445 & 0.250 & 0.439 & 0.154 & \textbf{0.524} & \textbf{0.336} \\
\midrule
\rowcolor{oursbg}[0pt][0pt] \multicolumn{10}{@{}l@{}}{\textcolor{oursblue!75!black}{\textbf{Sensor shifted 3\,m left}}}\\
\texttt{10072140} & 52 & 2173 & 290 & 0.529 & 0.276 & 0.600 & 0.079 & \textbf{0.672} & \textbf{0.424} \\
\texttt{10500357} & 52 & 2237 & 123 & 0.215 & 0.054 & 0.260 & 0.066 & \textbf{0.286} & \textbf{0.094} \\
\texttt{10676267} & 52 & 798 & 48 & 0.115 & 0.002 & 0.151 & 0.012 & \textbf{0.195} & \textbf{0.053} \\
\texttt{10830568} & 52 & 1419 & 97 & 0.153 & 0.196 & 0.307 & 0.426 & \textbf{0.425} & \textbf{0.589} \\
\texttt{13271285} & 52 & 1723 & 418 & 0.138 & 0.014 & 0.314 & 0.136 & \textbf{0.472} & \textbf{0.315} \\
\texttt{17761959} & 54 & 405 & 9 & 0.273 & \textbf{0.000} & 0.326 & \textbf{0.000} & \textbf{0.382} & \textbf{0.000} \\
\cmidrule(lr){1-10}
\textbf{Pooled} & 314 & 8755 & 985 & 0.247 & 0.053 & 0.341 & 0.120 & \textbf{0.434} & \textbf{0.266} \\
\bottomrule
\end{tabularx}
\end{table*}

\textbf{Detector.}
We use CenterPoint~\cite{centerpoint} as the LiDAR detector. The detector is trained on TOP first-return clouds from the 790 Waymo segments outside the reserved evaluation set. We sample every fifth frame as training data and train for 30 epochs. The detector receives point coordinates and intensity, with its additional elongation input fixed to zero. Inputs are cropped to $x,y\in[-75.2,75.2]$\,m and $z\in[-2,4]$\,m. All reconstruction results are evaluated using the same detector checkpoint in Waymo's vehicle coordinate frame.

\textbf{Detection metrics.}
We report vehicle and pedestrian LEVEL\_1 AP and APH, where APH weights detections by heading accuracy. Predictions use a confidence threshold of $0.1$ and rotated BEV non-maximum suppression at IoU $0.7$, with at most 500 output boxes per scan. The Waymo evaluator uses Hungarian matching of 3D boxes at IoU $0.7$ for vehicles and $0.5$ for pedestrians, with a target of 11 score thresholds. For the overall scores, detections from all six segments are pooled before evaluation. Reconstruction-input frames and held-out frames are scored separately. For the teaser, we use all 314 recorded-trajectory frames and average the vehicle and pedestrian LEVEL\_1 AP values.

\textbf{Shifted-view evaluation.}
No recorded scan is available at the shifted sensor position. We evaluate detections against the original boxes, keeping their original point counts and difficulty labels. The scores therefore measure how well the reconstructed scene supports detection of the original annotated objects after moving the sensor. %

\section{Per-scene detection results}\label{sec:supp-scene-detection}
We compare NuRec, LiDAR-GS, and \ours{} separately on each of the six Waymo scenes in \cref{tab:supp-scene-detection}, using the detector and LEVEL\_1 AP settings in \cref{sec:supp-detection}. Each scene has four held-out queries on the recorded trajectory and 52 or 54 queries with the sensor shifted $3$\,m left. The table also reports the number of annotated vehicle and pedestrian box instances; an object appearing in multiple frames is counted once per frame.

\textbf{Held-out frames.}
\ours{} achieves the highest pooled vehicle and pedestrian AP among the three methods, at $0.524$ and $0.336$. The comparison varies across scenes: its vehicle AP exceeds NuRec in three of six scenes and LiDAR-GS in five of six. Pedestrian AP exceeds LiDAR-GS in all five scenes with pedestrian annotations. %

\textbf{Shifted views.}
With the sensor shifted $3$\,m left, \ours{} has higher vehicle AP than both NuRec and LiDAR-GS in all six scenes. Its pedestrian AP is higher in five scenes; all three methods score zero in \texttt{17761959}, which contains nine pedestrian instances. These results show that our detection advantage is consistent across scenes under novel view evaluations.

\section{Intensity and return evaluation}\label{sec:supp-signal}
\textbf{Projection.}
We evaluate the same 60 PAI and 24 Waymo held-out queries on the recorded trajectories used for geometric reconstruction. For each query, we transform both point clouds to the sensor coordinate frame at sweep end and project them onto the same angular grid: $128\times3600$ pixels for PAI and $64\times2640$ for Waymo. If multiple points occupy a pixel, we keep the nearest return and its intensity.

\textbf{Metrics.}
Let $G$ and $P$ denote the sets of occupied pixels in the recorded and reconstructed panoramas. We compute intensity mean absolute error (MAE) at pixels in $G\cap P$, and define return F1 and recall as
\begin{equation}
F_1=\frac{2|G\cap P|}{|G|+|P|},\qquad
\mathrm{Recall}=\frac{|G\cap P|}{|G|}.
\end{equation}

\begin{table}[!ht]\centering
\caption{\textbf{Intensity and return agreement on held-out scans.} Intensity MAE compares pixels occupied in both projected clouds. Return F1 and recall compare the occupancy masks. Scores are averaged within clips and then across six clips. Bold marks the best value in each dataset.}\label{tab:ray}
\small\setlength{\tabcolsep}{2pt}\renewcommand{\arraystretch}{1.08}
\begin{tabularx}{\columnwidth}{@{}lRRR@{}}
\toprule
Method & \shortstack{Intensity\\MAE$\downarrow$} & \shortstack{Return\\F1$\uparrow$} & \shortstack{Return\\recall$\uparrow$} \\
\midrule
\rowcolor{refgray}[0pt][0pt] \multicolumn{4}{@{}l@{}}{\textbf{PAI: 60 held-out scans}}\\
NuRec & 0.0286 & \textbf{0.792} & 0.877 \\
LiDAR-GS & \textbf{0.0241} & 0.690 & 0.731 \\
SplatAD & 0.0375 & 0.760 & 0.811 \\
LiDAR-EVS & 0.0391 & 0.770 & \textbf{0.905} \\
\rowcolor{oursbg} \textcolor{oursblue}{\textbf{\ours}} & 0.0300 & 0.654 & 0.591 \\
\midrule
\rowcolor{refgray}[0pt][0pt] \multicolumn{4}{@{}l@{}}{\textbf{Waymo: 24 held-out scans}}\\
NuRec & \textbf{0.0264} & 0.929 & 0.932 \\
LiDAR-GS & \textbf{0.0264} & 0.852 & 0.817 \\
SplatAD & 0.0316 & \textbf{0.942} & \textbf{0.976} \\
LiDAR-EVS & 0.0328 & 0.937 & 0.974 \\
\rowcolor{oursbg} \textcolor{oursblue}{\textbf{\ours}} & 0.0390 & 0.839 & 0.785 \\
\bottomrule
\end{tabularx}
\end{table}

\textbf{Quantitative comparison.}
On the 24 Waymo held-out queries, \ours{} has higher intensity MAE ($0.0390$ versus $0.0264$) and lower return F1 ($0.839$ versus $0.942$) than the best per-scene results (\cref{tab:ray}). Yet its pooled vehicle and pedestrian AP reaches $0.524$ and $0.336$, exceeding NuRec and LiDAR-GS (\cref{tab:supp-scene-detection}). These metrics assess different properties: reconstruction scores aggregate point or pixel agreement across the scan, including background surfaces, whereas detection AP evaluates recognition and localization of annotated objects. Missing background returns or intensity differences can worsen reconstruction scores while leaving object detections unchanged. Conversely, a local distortion of an object's shape or position can affect its box match while contributing little to a scan-level score. Prior LiDAR simulation studies also use downstream perception to complement reconstruction evaluation~\cite{dynfl,lidargs}. In our pipeline, measurement-anchored Gaussian prediction constrains position changes around recorded surfaces, while sampling-aware Gaussian scales and multi-frame consolidation maintain local support and reduce redundant predictions. The main paper visualizations show preserved nearby vehicle shapes and well-aligned detection boxes alongside sparser distant coverage. These observations help explain why lower overall reconstruction fidelity can coexist with higher detection AP. In summary, \ours{} combines minute-scale reconstruction with strong detection performance, enabling faster construction and updating of LiDAR simulation assets for autonomous driving.

\end{document}